\documentclass[runningheads]{llncs}
\usepackage{graphicx}
\usepackage{comment}
\usepackage{amsmath,amssymb} % define this before the line numbering.
\usepackage{color}
\usepackage{subfigure}
\usepackage{multirow}
\usepackage{booktabs}
\usepackage{float}

\begin{document}
% \linenumbers
\pagestyle{headings}
\mainmatter
\def\ECCVSubNumber{2047}  % Insert your submission number here

\title{XSepConv: Extremely Separated Convolution} % Replace with your title

\begin{comment}
\titlerunning{ECCV-20 submission ID \ECCVSubNumber} 
\authorrunning{ECCV-20 submission ID \ECCVSubNumber} 
\author{Anonymous ECCV submission}
\institute{Paper ID \ECCVSubNumber}
\end{comment}
%******************

%\begin{comment}
\titlerunning{XSepConv}
\author{Jiarong Chen\inst{1}\and
Zongqing Lu\inst{1} \and
Jing-Hao Xue\inst{2} \and
Qingmin Liao\inst{*1}}
\authorrunning{Chen et al.}
\institute{Tsinghua University\\
\email{cjr18@mails.tsinghua.edu.cn,luzq@sz.tsinghua.edu.cn,\\liaoqm@tsinghua.edu.cn}\\
\and
University College London\\
\email{jinghao.xue@ucl.ac.uk}\\
* Corresponding author}
%\end{comment}
%******************
\maketitle

\begin{abstract}
Depthwise convolution has gradually become an indispensable operation for modern efficient neural networks and larger kernel sizes ($\ge5$) have been applied to it recently. In this paper, we propose a novel extremely separated convolutional block (XSepConv), which fuses spatially separable convolutions into depthwise convolution to further reduce both the computational cost and parameter size of large kernels. Furthermore, an extra $2\times2$ depthwise convolution coupled with improved symmetric padding strategy is employed to compensate for the side effect brought by spatially separable convolutions. XSepConv is designed to be an efficient alternative to vanilla depthwise convolution with large kernel sizes. To verify this, we use XSepConv for the state-of-the-art architecture MobileNetV3-Small and carry out extensive experiments on four highly competitive benchmark datasets (CIFAR-10, CIFAR-100, SVHN and Tiny-ImageNet) to demonstrate that XSepConv can indeed strike a better trade-off between accuracy and efficiency.
\end{abstract}

\section{Introduction}

Convolutional neural networks (CNNs) are becoming increasingly ubiquitous in numerous computer vision tasks, such as object detection and image classification, due to their more and more outstanding performance over time. Consequently, it has stimulated the desire to deploy these top-performing CNNs on resource-constrained platforms, e.g., on mobile phones, drones, self-driving cars, robots and Internet-of-Things (IOT) devices. However, most top-performing CNNs are in need of tremendous computational resources, severely impeding their practical deployment on these devices with constrained computing power.

Regarding the issue mentioned above, a lot of research work has been dedicated to the design of efficient CNN architectures, leading to the emergence of a variety of architectures with outstanding performance in terms of accuracy and efficiency trade-off, including Xception \cite{chollet2017xception}, MobileNets \cite{howard2017mobilenets,sandler2018mobilenetv2,howard2019searching}, ShuffleNets \cite{zhang2018shufflenet,ma2018shufflenet}, NASNet \cite{zoph2018learning}, MnasNet \cite{tan2019mnasnet}, EfficientNet \cite{tan2019efficientnet} and IGCV family \cite{zhang2017interleaved,xie2018interleaved,sun2018igcv3}, to name a few. Among these top-performing architectures, the CNNs built upon depthwise convolution, which are represented by the family of MobileNets \cite{howard2017mobilenets,sandler2018mobilenetv2,howard2019searching}, are increasingly becoming the mainstream attributing to their better trade-off between accuracy and efficiency.

A standard depthwise convolution can be viewed as a special case of group convolution, where the number of groups is equal to the number of channels. With an input tensor of $N$ channels, it contains $N$ kernels, each of which is independently applied to one channel of the input tensor, thereby reducing both the computational cost and parameter count by a factor of $N$. Since its extensive adoption in the milestone efficient CNN architecture MobileNets \cite{howard2017mobilenets}, depthwise convolution has attracted a lot of research interest in utilizing it to design efficient CNNs. In the early stage, the design of efficient CNNs is mainly through manual effort, and the major research effort is devoted to devising the basic building blocks as well as overall architectures of neural networks. As a result, the kernel size of depthwise convolution is simply specified as 3 in most cases \cite{chollet2017xception,howard2017mobilenets,sandler2018mobilenetv2,zhang2018shufflenet,ma2018shufflenet,xie2018interleaved,liu2018darts}. Recently, neural architecture search (NAS) has been exploited to automatically design efficient CNNs, leading to the adoption of larger depthwise convolutional kernel sizes \cite{zoph2018learning,pham2018efficient,liu2018progressive,real2019regularized,cai2018proxylessnas,wu2019fbnet,tan2019mnasnet,howard2019searching,tan2019efficientnet,tan2019mixconv}. Moreover, by adopting symmetric padding \cite{wu2019convolution}, the conventional convolution with even-sized kernels can also achieve competitive accuracy compared with depthwise convolution.

Recognizing that the CNNs built upon depthwise convolution mainly focus on exploration in the dimension of channels to reach the aim of reduction in computational cost, potentially more reduction can be achieved through further decomposition in the spatial dimension of orthogonal space, especially for large depthwise convolutional kernels. Spatially separable convolutions can be viewed as a decomposition at spatial level where a width-wise convolution followed by a height-wise convolution is composed to be an approximate replacement of original two-dimensional spatial convolution, thereby shrinking the computational complexity. However, it will suffer significant information loss if spatially separable convolutions are directly employed in network architectures \cite{jin2014flattened}.

\begin{figure}[t]
  \centering
  \subfigure[]{
    \label{fig:XSepConv:a} %% label for first subfigure
    \includegraphics[width=0.48\linewidth]{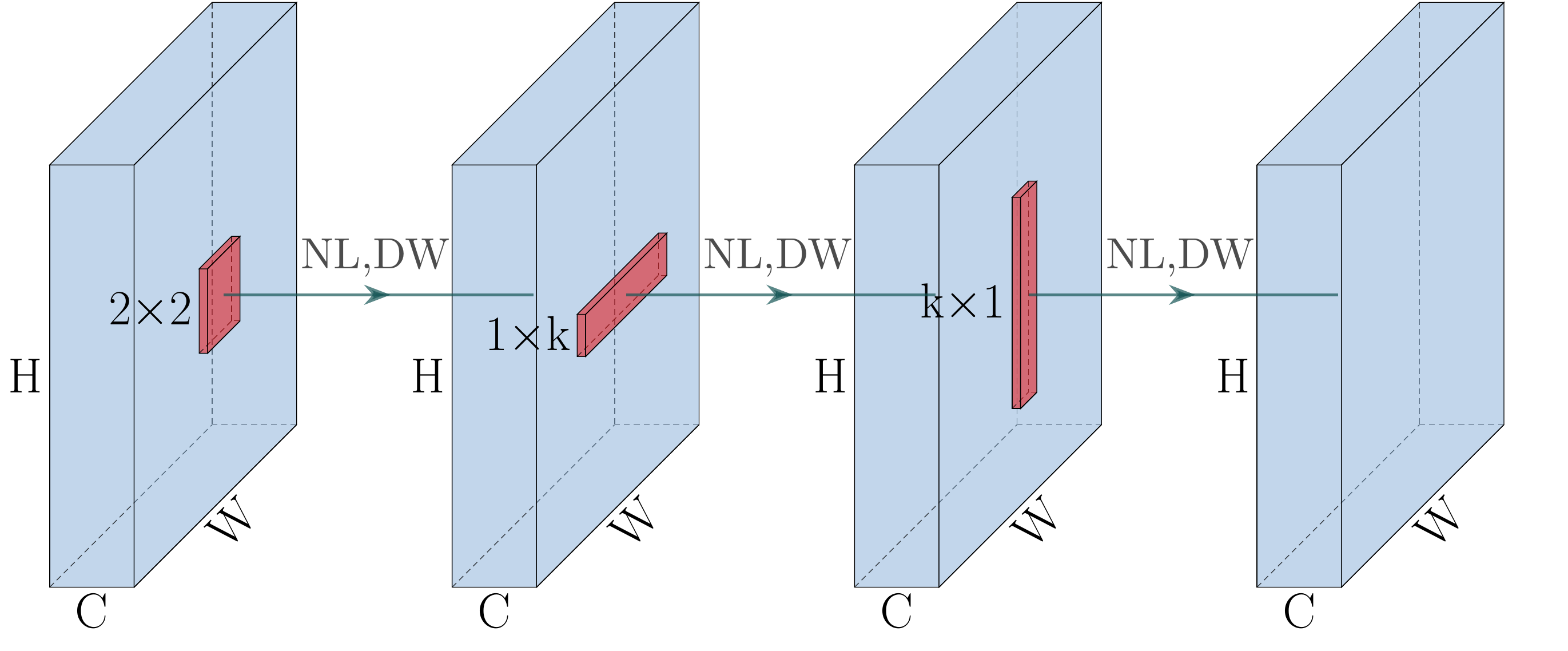}}
  \subfigure[]{
    \label{fig:XSepConv:b} %% label for second subfigure
    \includegraphics[width=0.48\linewidth]{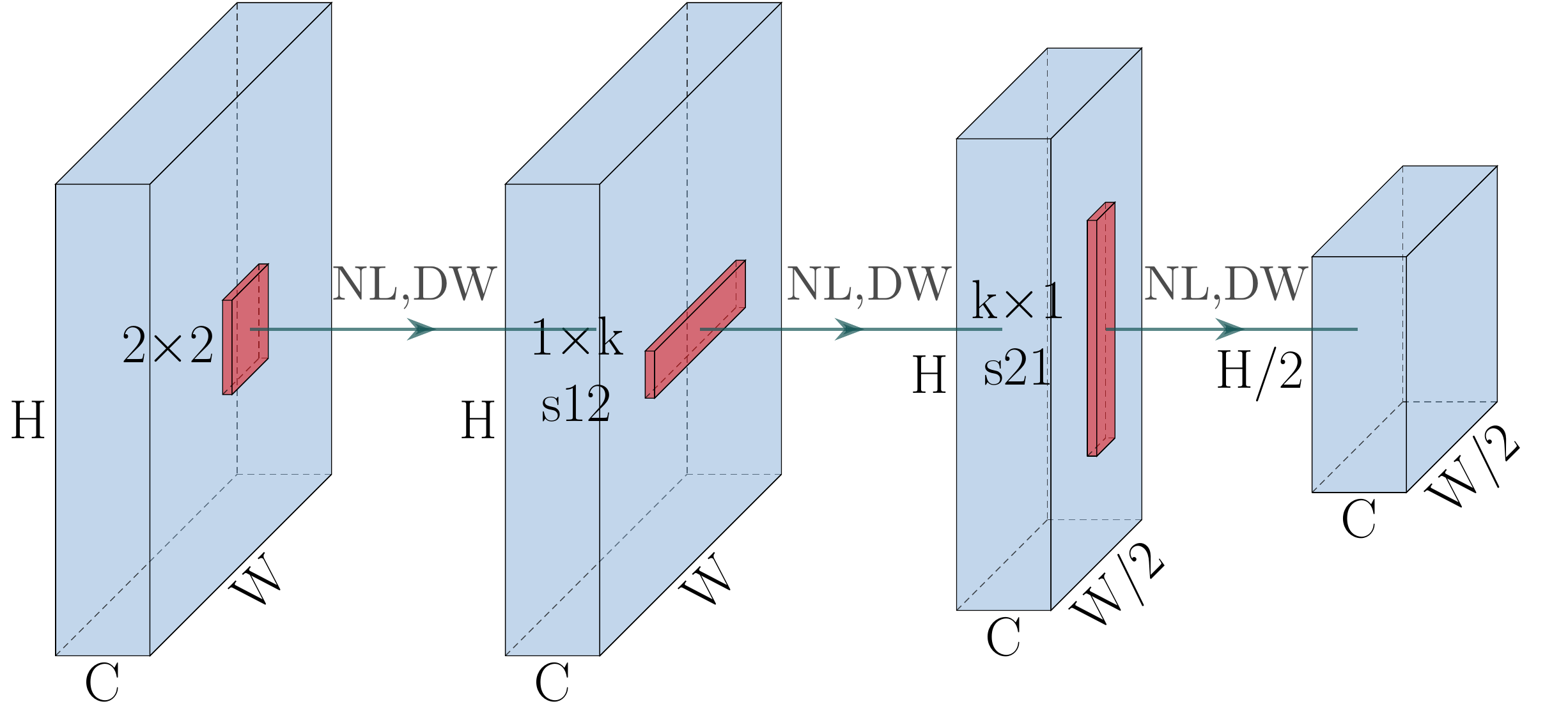}}
  \caption{(a) the basic XSepConv block. (b) the XSepConv block for spatial down sampling (2$\times$). \textbf{NL}: nonlinearity. \textbf{DW}: depthwise convolution. \textbf{s12}: stride=(1,2). \textbf{s21}: stride=(2,1). (Better viewed in color)}
  \label{fig:XSepConv} %% label for entire figure
\end{figure}

\begin{figure}[t]
  \centering
  \subfigure[]{
    \label{fig:padding:a} %% label for first subfigure
    \includegraphics[width=0.48\linewidth]{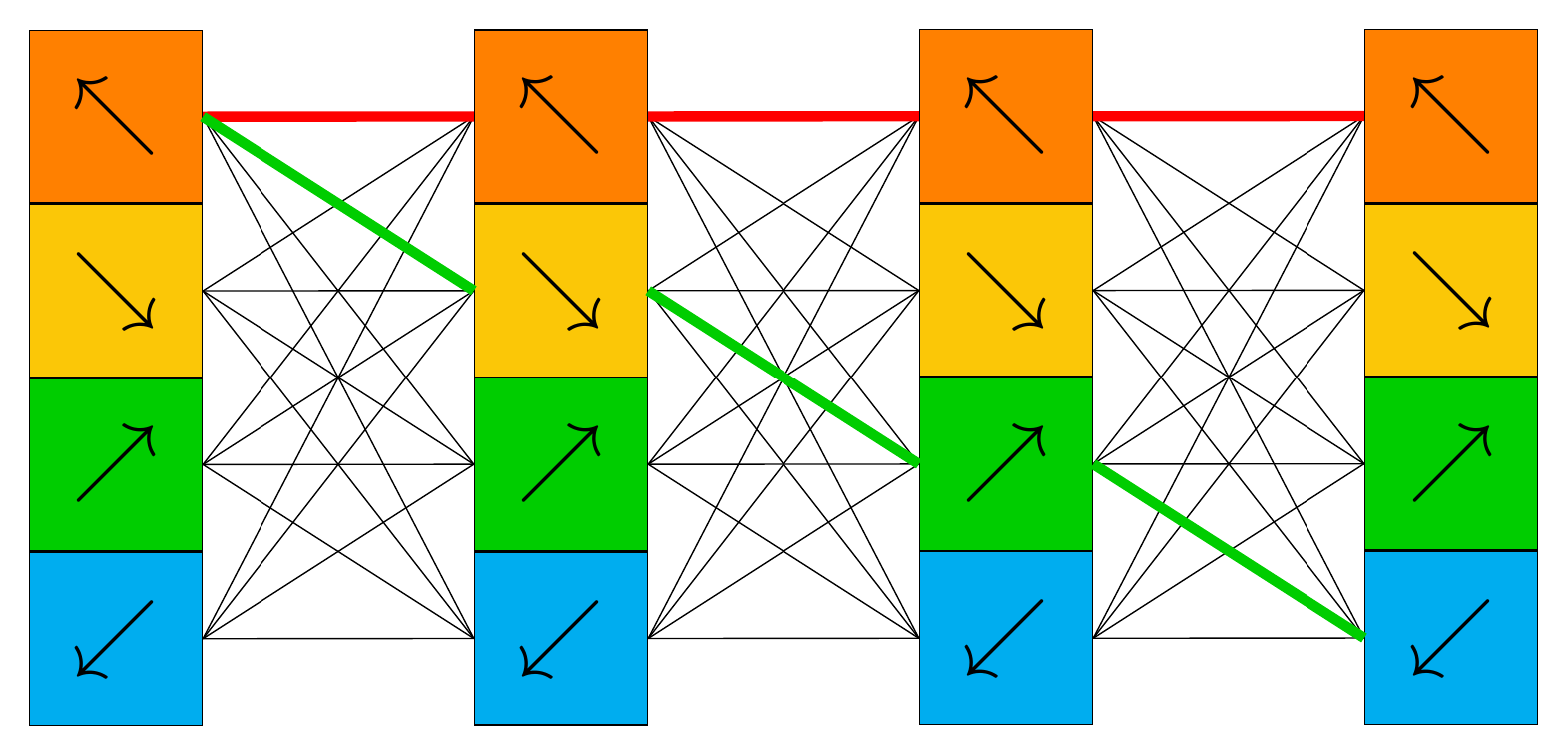}}
  \subfigure[]{
    \label{fig:padding:b} %% label for second subfigure
    \includegraphics[width=0.48\linewidth]{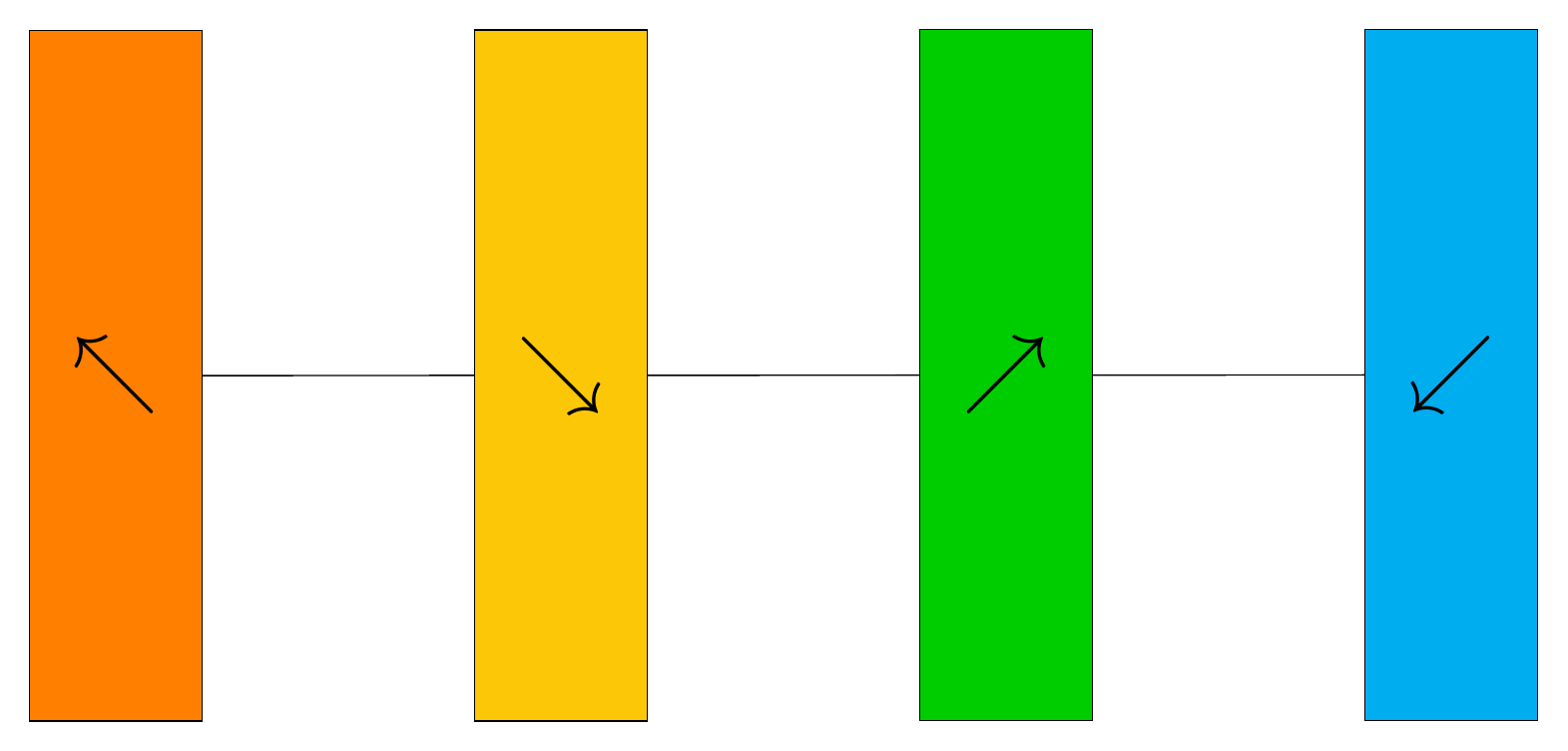}}
  \caption{Two symmetric padding strategies performed on four consecutive layers. (a) Symmetric padding strategy proposed in \cite{wu2019convolution}, different colors represent different groups in a single layer. (b) Our proposed improved symmetric padding strategy, different colors represent different layers. The direction of the arrow indicates the direction of position offset, e.g., the arrow in the left-top direction illustrates the shift in the left-top direction, which is caused by padding one more zero on the right and bottom sides of feature maps. The red line and green line represent ``bad'' and ``good'' information flow paths respectively, please refer to Section~\ref{padding} for details. (Best viewed in color)}
  \label{fig:padding} %% label for entire figure
\end{figure}

In this paper, we propose an extremely separated convolutional block, dubbed XSepConv, which mixes depthwise convolution with spatially separable convolutions to form spatially separated depthwise convolutions, further reducing the parameter size and computational burden of large depthwise convolutional kernels. Considering that spatially separable convolutions lack sufficient ability to capture information except in vertical and horizontal directions, additional operations are required to capture information in other directions (e.g. diagonal direction) to avoid significant information loss. Here we employ a simple but effective operation, $2\times2$ depthwise convolution with improved symmetric padding strategy, to compensate for the above-mentioned side effect to a certain degree. Fig.~\ref{fig:XSepConv:a} shows the basic structure of XSepConv, which is composed of $2\times2$ depthwise convolution followed by spatially separated depthwise convolutions. For spatial downsampling, the structure is illustrated in Fig.~\ref{fig:XSepConv:b}, which is divided into two downsampling phases, width-wise and height-wise, to preserve as much information as possible during downsampling. Fig.~\ref{fig:padding:b} illustrates our proposed improved symmetric padding strategy, which performs symmetric padding within four successive even-sized convolution layers instead of a single even-sized convolution layer \cite{wu2019convolution} as shown in Fig.~\ref{fig:padding:a}. Extensive experiments on four highly competitive benchmark datasets (CIFAR-10, CIFAR-100, SVHN and Tiny-ImageNet) show that simply replacing large depthwise convolution kernels with XSepConv in MobileNetV3-Small achieves a solid accuracy improvement with fewer parameters and FLOPs (floating point operations). Therefore, we demonstrate that XSepConv is a simple while efficient replacement of vanilla depthwise convolution with large kernel sizes, which can strike a better trade-off between accuracy and efficiency.

\section{Related Work}

\subsection{Efficient CNNs}

In recent years, with the increasing demand for deploying CNNs on resource-constrained platforms, a series of studies have been conducted in designing efficient CNNs to strike an optimal trade-off between accuracy and efficiency. SqueezeNet \cite{iandola2016squeezenet} extensively utilizes $1\times1$ convolution in the Fire module and achieves AlexNet-level accuracy with 50$\times$ fewer parameters. Xception \cite{chollet2017xception} obtains the performance improvement due to a more efficient use of model parameters by adopting depthwise separable convolutions. MobileNetV1 \cite{howard2017mobilenets} is built on depthwise separable convolutions composed of a depthwise convolution followed by a pointwise convolution, greatly improving the computational efficiency. MobileNetV2 \cite{sandler2018mobilenetv2} introduces resource-efficient inverted residuals with linear bottlenecks. Based on MobileNetV2, MobileNetV3 \cite{howard2019searching} integrates squeeze-and-excitation module \cite{hu2018squeeze} into the bottleneck structure and applies modified swish nonlinearities, subsequently these architecture advances are blended with hardware-aware neural architecture search to build efficient models. ShuffleNets \cite{zhang2018shufflenet,ma2018shufflenet} further reduce computational cost by utilizing pointwise group convolution and channel shuffle. CondenseNet \cite{huang2018condensenet} combines dense connectivity with learned group convolution to promote feature re-use while eliminating redundant connections. IGCNets\cite{zhang2017interleaved,xie2018interleaved,sun2018igcv3} propose interleaved group convolutions which are efficient in terms of parameter and computation. ShiftNet \cite{wu2018shift} introduces shift operations to replace expensive spatial convolutions.

Recently, neural architecture search has been leveraged to automate the model design process, thus spawning a series of efficient models, such as NASNet \cite{zoph2018learning}, PNASNet \cite{liu2018progressive}, AmoebaNet \cite{real2019regularized}, FBNet \cite{wu2019fbnet}, MnasNet \cite{tan2019mnasnet}, EfficientNet \cite{tan2019efficientnet} and MixNets \cite{tan2019mixconv}.

\subsection{Separable Convolutions}

Depthwise separable convolutions, first introduced in \cite{sifre2014rigid}, factorize a standard convolution into a depthwise convolution followed by a pointwise convolution and can be viewed as a decomposition along the channel domain. Since Xception \cite{chollet2017xception} and MobileNets \cite{howard2017mobilenets} which are built upon depthwise separable convolutions achieved great success in terms of accuracy and efficiency trade-off, depthwise separable convolutions have attracted a lot of research efforts. Therefore, a variety of efficient models have emerged, from hand-crafted models \cite{chollet2017xception,howard2017mobilenets,sandler2018mobilenetv2,zhang2018shufflenet,ma2018shufflenet,xie2018interleaved} to models found by NAS \cite{liu2018darts,zoph2018learning,pham2018efficient,liu2018progressive,real2019regularized,cai2018proxylessnas,wu2019fbnet,tan2019mnasnet,howard2019searching,tan2019efficientnet,tan2019mixconv}.

Compared with depthwise separable convolutions, spatially separable convolutions or called asymmetric convolutions, which can be regarded as a decomposition in the spatial dimension, have been applied earlier in CNNs \cite{mamalet2012simplifying}. Spatially separable convolutions factorize a traditional two-dimensional ($k\times k$) convolution into a width-wise ($k\times1$) convolution and a height-wise ($1\times k$) convolution to reduce the number of parameters and the computation. However, it will result in significant information loss if the separation is directly applied to filters \cite{jin2014flattened}. Several methods have been proposed to tackle this problem, e.g.,  by deriving an appropriate low-rank approximation using SVD \cite{denton2014exploiting}, by minimizing the $L_{2}$ reconstruction error \cite{jaderberg2014speeding}, and by applying structural constraints \cite{jin2014flattened}. In addition, spatially separable convolutions are widely applied as a building unit in CNNs such as Inception-v3 \cite{szegedy2016rethinking} and Inception-v4 \cite{szegedy2017inception}.

\subsection{Kernel Sizes}

In the hand-crafted efficient models, the most commonly used kernel size of depthwise convolution is 3 \cite{chollet2017xception,howard2017mobilenets,sandler2018mobilenetv2,zhang2018shufflenet,ma2018shufflenet,xie2018interleaved}. Recently, by adding depthwise convolution with various kernel sizes into the search space, larger kernels are applied to CNNs found by NAS \cite{zoph2018learning,pham2018efficient,liu2018progressive,real2019regularized,cai2018proxylessnas,wu2019fbnet,tan2019mnasnet,howard2019searching,tan2019efficientnet,tan2019mixconv}, e.g., $5\times5$ kernels in MobileNetV3 \cite{howard2019searching} and $7\times7$ kernels in ProxylessNAS \cite{cai2018proxylessnas}, which have shown their potential abilities to improve accuracy and efficiency. Besides, direct implementation of even-sized kernels (2$\times$2, 4$\times$4) will encounter performance degradation due to the shift problem, and applying symmetric padding can eliminate the problem, thus improving the generalization abilities of even-sized kernels \cite{wu2019convolution}.

\section{XSepConv}

The main idea of XSepConv is to utilize spatially separable convolutions to reduce the parameter size and computational complexity of large depthwise convolution kernels, and an extra $2\times2$ depthwise convolution with improved symmetric padding strategy is used to compensate the side effect brought by spatially separable convolutions. In this section, we will describe the advantages of XSepConv and the improved symmetric padding strategy.

\subsection{Wider and Deeper}

A standard depthwise convolutional kernel $\mathbf{W}\in \mathbb{R}^{k\times k\times c}$ performs a 2D convolution on each individual channel of the input tensor $\mathbf{X}\in \mathbb{R}^{h\times w\times c}$, and produces the output tensor $\mathbf{Y}\in \mathbb{R}^{h\times w\times c}$ with the same shape, where $k$ denotes the kernel size, $h$ denotes the spatial height, $w$ denotes the spatial width and $c$ denotes the number of channels. It can be formulated as
\begin{align}
  \mathbf{Y}_{x,y,z}=\sum_{-\frac{k}{2}\leq i,j\leq\frac{k}{2}}\mathbf{W}_{i,j,z}\mathbf{X}_{x+i,y+j,z} .
\end{align}

The computational cost of the vanilla depthwise convolution is $k^{2}hwc$ and the parameter size is $k^{2}c$. It can be obviously seen that the computational cost and parameter size will increase in quadratic as the kernel size increases. In addition, the receptive field size of the depthwise convolution is equal to its kernel size $k$.

As shown in Fig.~\ref{fig:XSepConv:a}, XSepConv consists of three parts: $2\times2$ depthwise convolutional kernel $\hat{\mathbf{W}}'\in\mathbb{R}^{2\times 2\times c}$, $1\times k$ depthwise convolutional kernel $\hat{\mathbf{W}}''\in\mathbb{R}^{1\times k\times c}$ and $k\times1$ depthwise convolutional kernel $\hat{\mathbf{W}}'''\in \mathbb{R}^{k\times 1\times c}$. The $1\times k$ and $k\times1$ depthwise convolutions together form what we call spatially separated depthwise convolutions, and the $2\times2$ depthwise convolution plays an important role in capturing information that may be missed by spatially separated depthwise convolutions due to their inherent structural defects. Taking padding one more zero on the right and bottom sides of the input tensor $\hat{\mathbf{X}}\in\mathbb{R}^{h\times w\times c}$ before $2\times2$ depthwise convolution as an example, the output tensor $\hat{\mathbf{Y}}\in\mathbb{R}^{h\times w\times c}$ is calculated as
\begin{align}
  \hat{\mathbf{Y}}_{x,y,z}=\sum_{-\frac{k}{2}\leq m\leq\frac{k}{2}}\sum_{-\frac{k}{2}\leq n\leq\frac{k}{2}}\sum_{0\leq i,j\leq 1}\hat{\mathbf{W}}_{i,j,z}'\hat{\mathbf{W}}_{0,n,z}''\hat{\mathbf{W}}_{m,0,z}'''\hat{\mathbf{X}}_{x+i+m,y+j+n,z} . 
  \label{eq:xsepconv}
\end{align}

XSepConv has the computational cost of $4hwc+2khwc$ and the parameter size of $4c+2kc$. Then the ratio of the computational cost (and also parameter size) of XSepConv to the vanilla depthwise convolution is calculated as
\begin{align}
    \frac{4hwc+2khwc}{k^{2}hwc}=\frac{4+2k}{k^{2}} .
\end{align}

Since we are focusing on replacing large depthwise convolutional kernel ($k\ge5$) with XSepConv, the ratio is consistently less than 1, which means that XSepConv requires less computation and fewer parameters. For instance, when $k=5$, XSepConv can save 44\% of the computational cost and parameter size, and the reduction will be greater as kernel size increases. More than that, as described in Section~\ref{experiments}, XSepConv achieves better accuracy than vanilla depthwise convolution, which we believe is because XSepConv is wider and deeper.

From Eq.~\ref{eq:xsepconv}, we can see that the receptive field size of XSepConv is $k+1$, thus XSepConv has a wider receptive field. In effect, the widening comes from the $2\times2$ depthwise convolution. Fig.~\ref{fig:XSepConv:a} shows the structure of XSepConv, which is composed of three consecutive layers, making it deeper. Moreover, all three layers are followed by batch normalization \cite{ioffe2015batch} and nonlinearity, so XSepConv can also increase the nonlinearities of the network, thereby enhancing its representation ability.

As for the downsampling layer, the structure of XSepConv is described in Fig.~\ref{fig:XSepConv:b}. The downsampling is split into two stages: first along the width direction with stride (1,2) and then along the height direction with stride (2,1). Similarly, the ratio of the computational complexity and parameter count of downsampling XSepConv to downsampling depthwise convolution is computed as
\begin{align}
    \frac{4hwc+khwc/2+khwc/4}{k^{2}hwc/4}=\frac{16+3k}{k^{2}} .
\end{align}

Since the ratio is less than 1 only when $k\ge7$, it is only recommended to use downsampling XSepConv to replace downsampling depthwise convolution with kernel size no less than 7.

\subsection{Improved Symmetric Padding Strategy}\label{padding}

When using even-sized convolutions, asymmetric padding, such as padding on the right and bottom sides only, is often used in order to maintain the size of feature maps, thus the activated values are shifted to the left-top corner of the spatial location, which is identified as the shift problem in \cite{wu2019convolution}. This problem limits the generalization abilities of even-sized kernels and a symmetric padding strategy which introduces symmetric padding within a single convolution layer is proposed in \cite{wu2019convolution} to eliminate the problem.

Fig.~\ref{fig:padding:a} illustrates the symmetric padding strategy proposed in \cite{wu2019convolution}, it focuses on introducing symmetry to the output of a single convolution layer. Through dividing the input tensor into four groups and padding in the left-top, right-bottom, left-bottom and right-top directions respectively, the location offset in the output feature maps of a single convolution layer is eliminated.

However, in most computer vision tasks such as image classification, the output of the last layer rather than the intermediate single layer is the most significant. As shown in Fig.~\ref{fig:padding:a}, the symmetric padding strategy of \cite{wu2019convolution} will encounter asymmetry at some outputs of the last layer. If the information flows along a path where all four shifted directions are different, such as the green line in Fig.~\ref{fig:padding:a}, the position offset will be eliminated in the final output. Otherwise, position offsets will accumulate in at least one direction, e.g., the red line in Fig.~\ref{fig:padding:a} indicates that the position offset in the left-top direction will accumulate in the final output, which will eventually squeeze features to the left-top corner of the spatial position and result in performance degradation.

Therefore, we propose an improved symmetric padding strategy as illustrated in Fig.~\ref{fig:padding:b}, which aims at the final output of the network instead of any intermediate single layer. We introduce a set of padding directions
\begin{align}
    \left\{\mathcal{D}_{1},\mathcal{D}_{2},\mathcal{D}_{3},\mathcal{D}_{0}\right\}
\end{align}

\noindent that in turn contains four directions: right-bottom, left-top, left-bottom and right-top. Let $N$ be the number of layers with even-sized kernels and $p(i)$ be the padding direction of the $i$-th layer with even-sized kernels. If $N$ is an integer multiple of 4, the symmetry of the final output is strictly achieved by specifying the padding directions:
\begin{align}
    p(i)=\mathcal{D}_{i\%4} ,
\end{align}

\noindent where $\%$ denotes the modulo operation. Even if $N$ is not an integer multiple of 4, we can also ensure that the final position offset does not exceed 1 pixel in any direction. Here the order of padding directions counts, for example, in case of $N\%4=2$, the two padding directions of the last two layers with even-sized kernels are opposite, so as to eliminate the shift as much as possible. In addition, if $N\%4=1$, the last layer with even-sized kernels can use the original symmetric padding strategy to achieve slightly better symmetry.

\section{Experiments}\label{experiments}

We conduct extensive experiments on four widely used image classification benchmark datasets: CIFAR-10 and CIFAR-100 \cite{krizhevsky2009learning}, SVHN \cite{netzer2011reading} and Tiny-ImageNet\footnote{\url{https://tiny-imagenet.herokuapp.com/}}, based on the state-of-the-art models MobileNetV3-Small \cite{howard2019searching}. Each experiment was repeated 5 times to eliminate the effect brought by random initialization. In addition, all the architectures are implemented by using PyTorch \cite{paszke2019pytorch}.

% We describe various datasets used in the experiments and training settings in Section~\ref{datasetsAndTraining}. Section~\ref{performance} compares the performance of XSepConv against vanilla depthwise convolution. Section~\ref{ablation} reports extensive ablation experiments to shed light on the impact of different structural designs.

\subsection{Datasets and Training Settings}\label{datasetsAndTraining}

\subsubsection{CIFAR:} The two CIFAR datasets \cite{krizhevsky2009learning}, i.e., CIFAR-10 and CIFAR-100, both contain 60,000 colored natural images with a size of $32\times32$, of which 50,000 images are for training and 10,000 images are for test. The major difference between CIFAR-10 and CIFAR-100 is that they consist of different numbers of classes. CIFAR-10 contains 10 classes, which means that each class consists of 5,000 training images and 1,000 test images. Similarly, CIFAR-100 contains 100 classes, each of which consists of 500 training images and 100 test images. We follow the most common data augmentation scheme \cite{he2016deep,huang2017densely}: first pad 4 zeros on each sides of the images and then randomly crop them to the size of $32\times32$, followed by randomly flipping the images horizontally. We finally normalize the images with the channel means and standard deviations.

\subsubsection{SVHN:} The Street View House Numbers (SVHN) \cite{netzer2011reading} dataset consists of 10 classes, each of which corresponding to a certain number between 0 and 9. It contains real-world colored digit images of resolution $32\times32$. There are 73,257 training images, 26,032 test images and 531,131 additional training images. We use both the training and additional data for training without any data augmentation \cite{huang2017densely}. Normalization with the channel means and standard deviations is also performed.

\subsubsection{Tiny-ImageNet:} The Tiny-ImageNet dataset, which consists of 200 classes drawn from 1,000 classes of ImageNet \cite{russakovsky2015imagenet},  is actually a subset of ImageNet dataset. There are 500 training images, 50 validation images and 50 test images per class. The images are resized to $64\times64$, making Tiny-ImageNet more difficult to learn due to the loss of detailed information during downsampling. We follow the data augmentation scheme for training: crop the image with the size no less than 8\% of the image area and the aspect ratio limited to the interval $\left[3/4,4/3\right]$ as in \cite{szegedy2015going}, resize to $56\times56$ and randomly flip the image horizontally. Normalization with the channel means and standard deviations is used in the end.

\subsubsection{Training Settings:} All networks are optimized using stochastic gradient descent (SGD) with momentum 0.9. We use batch normalization after every convolution layer, and the weight decay is set to 6e-5 and batch size to 128. Following \cite{he2019bag}, we use cosine learning rate decay and a gradual learning rate warmup strategy for the first 5 epochs. On CIFAR and SVHN we train for 400 and 20 epochs, respectively, with an initial learning rate of 0.35. For Tiny-ImageNet, the initial learning rate is set to 0.15 and we train models for 200 epochs.

\setlength{\tabcolsep}{4pt}
\begin{table}[t]
% \small
\begin{center}
\caption{Performance comparison of XSepConv with vanilla depthwise convolution on 4 datasets. ``Params'' refers to  the number of parameters}
\label{table:fourdatasets}
\begin{tabular}{ccccc}
\toprule\noalign{\smallskip}
Convolution & Datasets & FLOPs & Params & Top-1 accuracy (\%)\\
\noalign{\smallskip}
\midrule
\noalign{\smallskip}
DWConv  & \multirow{2}{*}{CIFAR-10} & 17.51M & 1.52M & 92.97\\
XSepConv &  & \textbf{16.71M} & \textbf{1.50M} & \textbf{93.24}\\
\midrule
\noalign{\smallskip}
DWConv  & \multirow{2}{*}{CIFAR-100} & 17.60M & 1.61M & 73.69\\
XSepConv &  & \textbf{16.80M} & \textbf{1.59M} & \textbf{74.02}\\
\midrule
\noalign{\smallskip}
DWConv  & \multirow{2}{*}{SVHN} & 17.51M & 1.52M & 97.92\\
XSepConv &  & \textbf{16.71M} & \textbf{1.50M} & \textbf{97.97}\\
\midrule
\noalign{\smallskip}
DWConv  & \multirow{2}{*}{Tiny-ImageNet} & 51.63M & 1.71M & 59.32\\
XSepConv &  & \textbf{49.18M} & \textbf{1.70M} & \textbf{59.82}\\
\bottomrule
\end{tabular}
\end{center}
\end{table}
\setlength{\tabcolsep}{1.4pt}

\subsection{XSepConv Performance}\label{performance}

\subsubsection{Trade-off between Accuracy and Efficiency:} To verify that XSepConv is an efficient replacement of vanilla depthwise convolution with large kernel, we evaluate its performance on 4 highly competitive image classification datasets with the state-of-the-art architecture MobileNetV3-Small \cite{howard2019searching}, which extensively utilizes $5\times5$ depthwise convolution. We re-implement the MobileNetV3-Small and replace the $5\times5$ depthwise convolutions in the middle stage of the network with our proposed XSepConv of $k=5$. In particular, we replace the $5\times5$ depthwise convolutions in the last stage (after the last downsampling layer) with XSepConv of $k=3$ to gain more computational reduction. All downsampling layers keep using the original depthwise convolutions because the kernel size is less than 7. In addition, all the training details are the same for fair comparison.

\begin{figure}[t]
  \centering
  \subfigure[CIFAR-10]{
    \label{fig:tradeoff:a} %% label for first subfigure
    \includegraphics[width=0.485\linewidth]{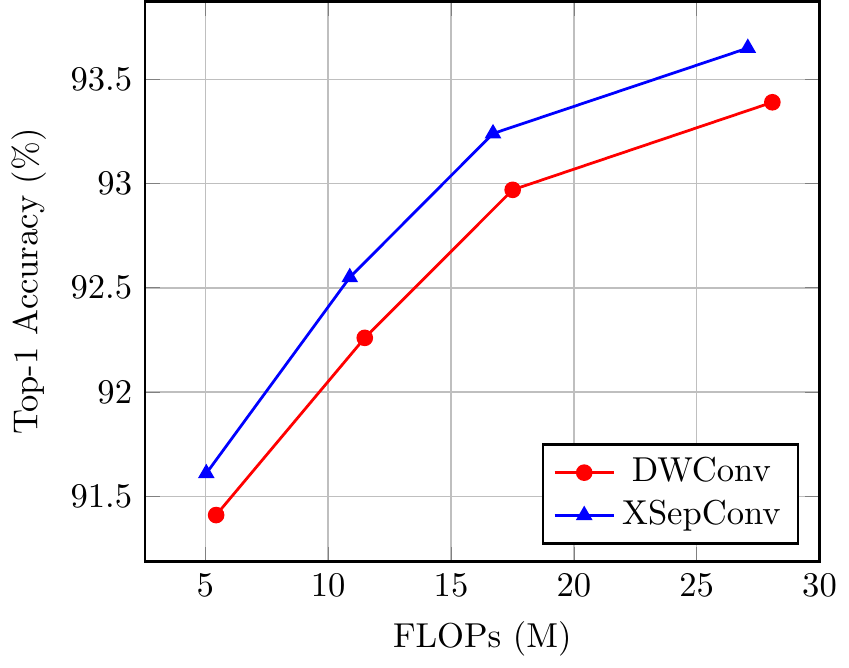}}
  \hfill
  \subfigure[CIFAR-100]{
    \label{fig:tradeoff:b} %% label for second subfigure
    \includegraphics[width=0.47\linewidth]{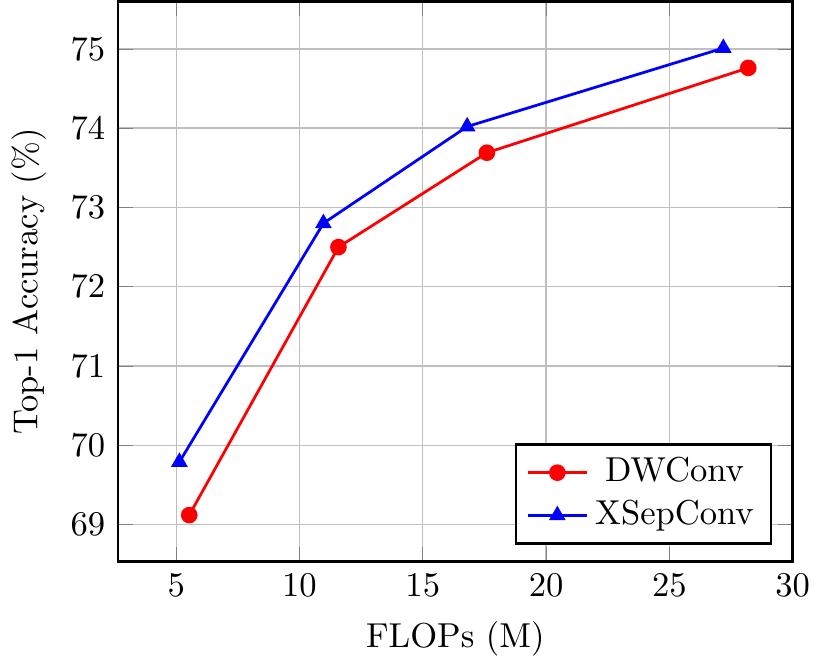}}
  \vfill
  \subfigure[SVHN]{
    \label{fig:tradeoff:c} %% label for third subfigure
    \includegraphics[width=0.485\linewidth]{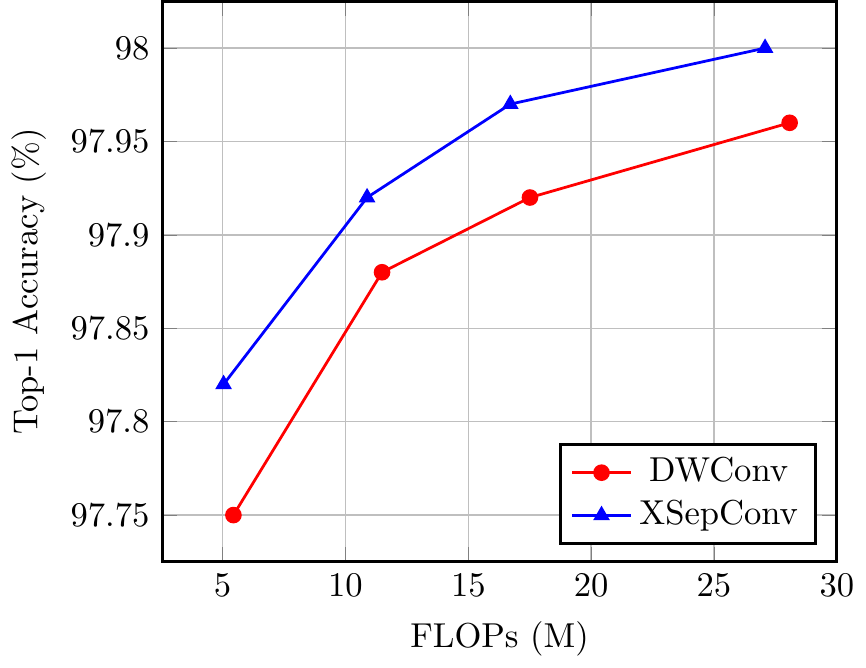}}
  \hfill
  \subfigure[Tiny-ImageNet]{
    \label{fig:tradeoff:d} %% label for fourth subfigure
    \includegraphics[width=0.455\linewidth]{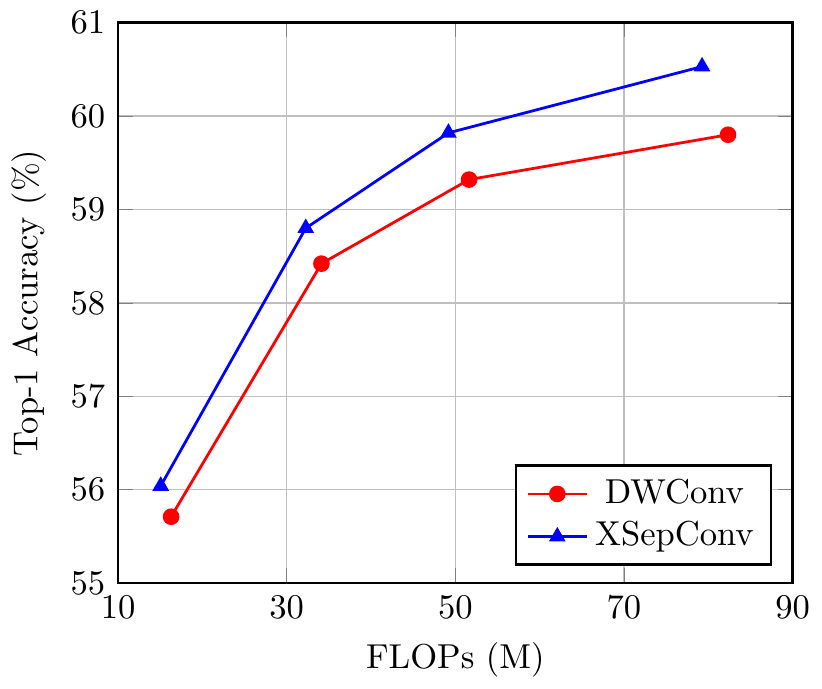}}
  \caption{Top-1 accuracy versus FLOPs on 4 datasets with various width multipliers. For each curve we use 4 different width multipliers 0.5, 0.75, 1.0 and 1.25. Curves further to the top left are more efficient in terms of accuracy and FLOPs trade-off. XSepConv outperforms its counterpart on all 4 datasets}
  \label{fig:tradeoff} %% label for entire figure
\end{figure}

Table~\ref{table:fourdatasets} shows the classification performance of XSepConv compared with vanilla depthwise convolution (DWConv). We compare the computational complexity (indicated by FLOPs), parameter size and top-1 accuracy on 4 datasets. It can be seen that XSepConv consistently obtains higher accuracy with fewer parameters and smaller computational complexity than vanilla depthwise convolution on all 4 datasets, illustrating that XSepConv is more efficient. For example, XSepConv acquires an additional performance improvement of 0.5\% while saving 4.75\% of the computational cost on Tiny-ImageNet. For more intuitive comparison, we provide the trade-off curves between top-1 accuracy and FLOPs on all 4 datasets in Fig.~\ref{fig:tradeoff}. Without bells and whistles, simply replacing ordinary depthwise convolution with XSepConv reliably improves the accuracy with fewer FLOPs under various computational complexity, indicating that XSepConv is able to achieve a more excellent trade-off between accuracy and efficiency.

% \begin{figure}[t]
%   \centering
%   \subfigure[]{
%     \label{fig:ks7:a} %% label for first subfigure
%     \includegraphics[width=0.48\linewidth]{fig/DS7-FLOPs.pdf}}
%   \hfill
%   \subfigure[]{
%     \label{fig:ks7:b} %% label for second subfigure
%     \includegraphics[width=0.48\linewidth]{fig/K7-FLOPs.pdf}}
%   \caption{Trade-off between top-1 accuracy and FLOPs on Tiny-ImageNet. (a) Comparison of downsampling XSepConv with downsampling depthwise convolution. (b) Comparison of XSepConv with depthwise convolution of kernel size 7}
%   \label{fig:ks7} %% label for entire figure
% \end{figure}
\begin{figure}[t]
\begin{minipage}[t]{0.48\linewidth}
\centering
\includegraphics[width=\linewidth]{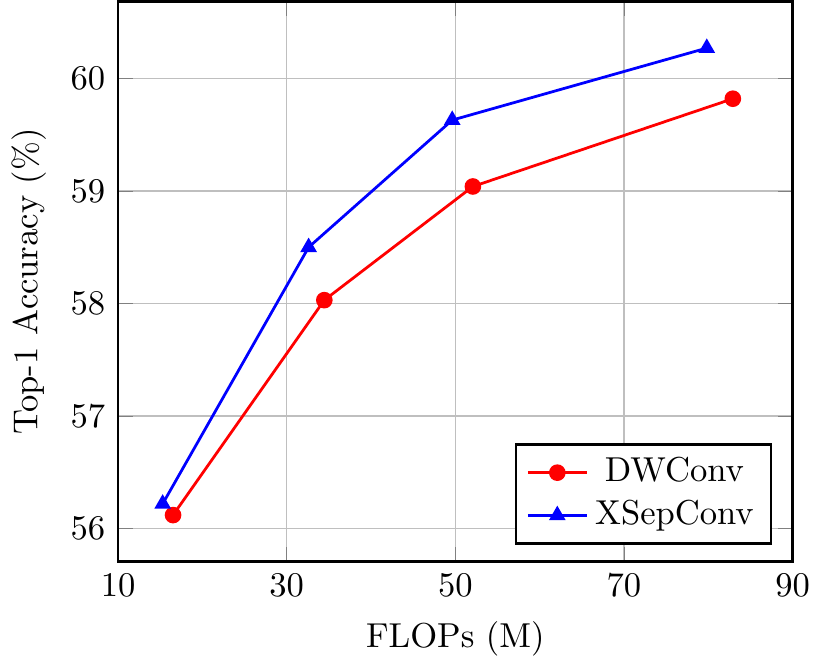}
\caption{Comparison of downsampling XSepConv with vanilla downsampling depthwise convolution of kernel size 7 on Tiny-ImageNet}
\label{fig:ds7}
\end{minipage}
\hfill
\begin{minipage}[t]{0.48\linewidth}
\centering
\includegraphics[width=\linewidth]{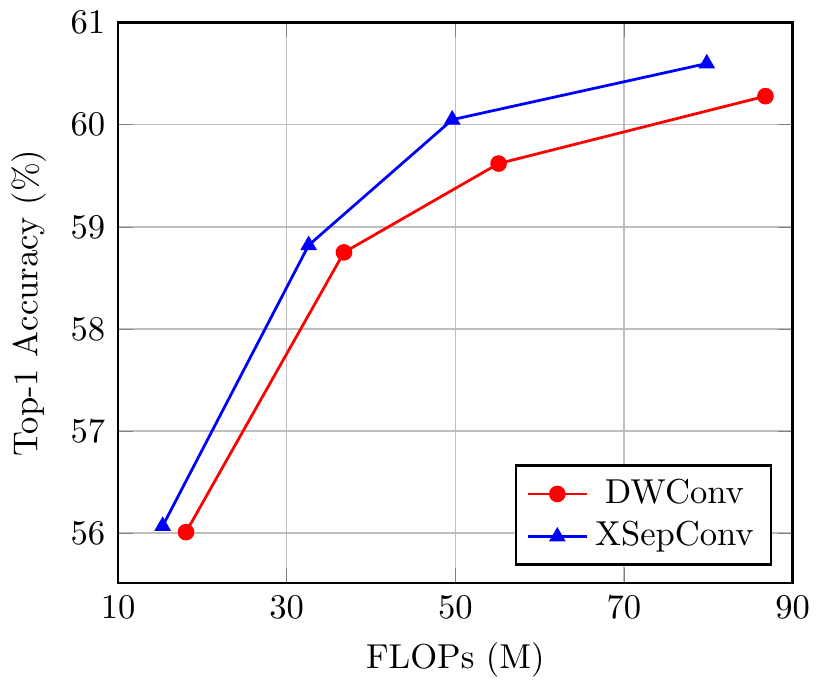}
\caption{Comparison of XSepConv with vanilla depthwise convolution of larger kernel size 7 on Tiny-ImageNet}
\label{fig:k7}
\end{minipage}
\end{figure}

\begin{figure}[!t]
  \centering
  \subfigure[]{
    \label{fig:BuildingBlock:a} %% label for first subfigure
    \includegraphics[width=0.21\linewidth]{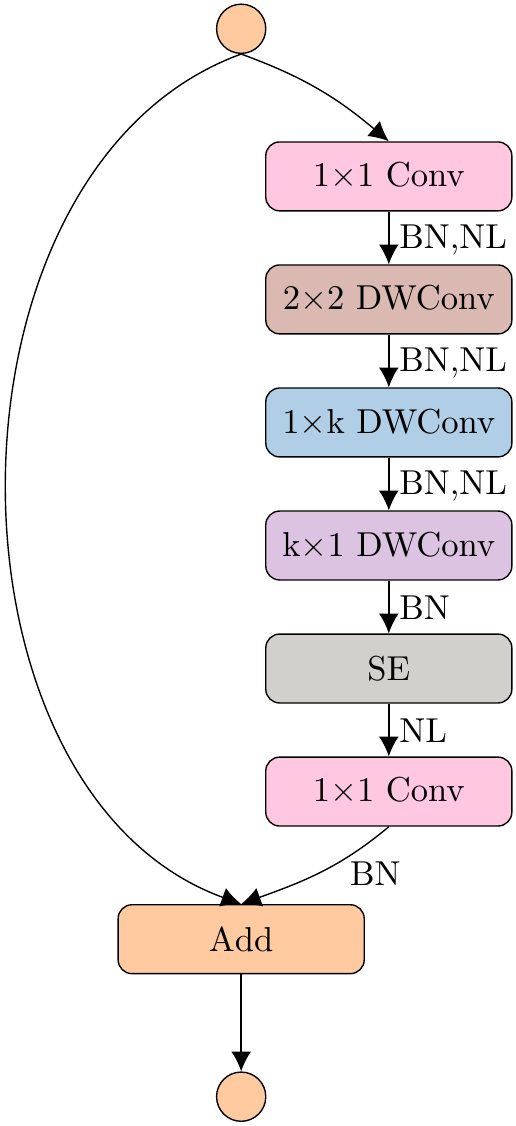}}
  \hfill
  \subfigure[]{
    \label{fig:BuildingBlock:b} %% label for second subfigure
    \includegraphics[width=0.25\linewidth]{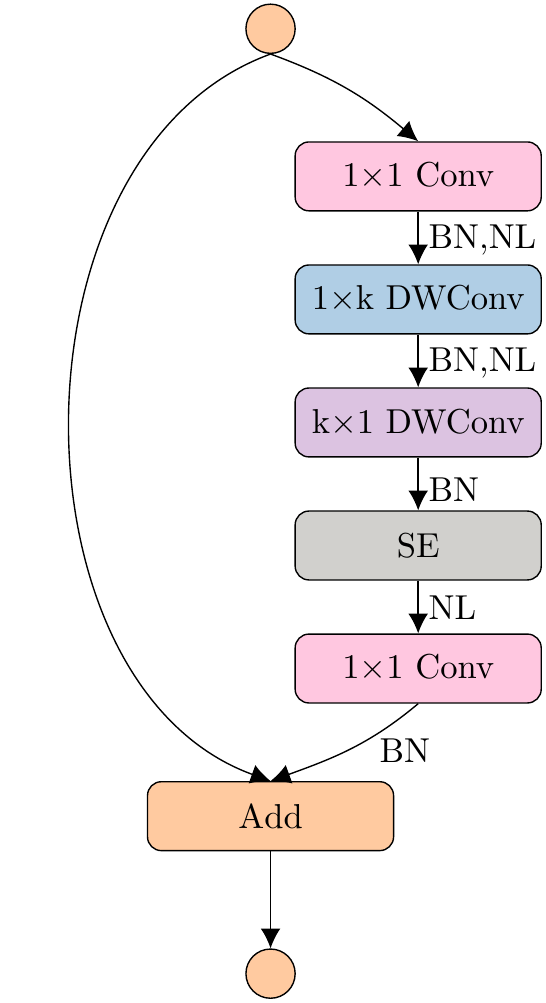}}
  \hfill
  \subfigure[]{
    \label{fig:BuildingBlock:c} %% label for second subfigure
    \includegraphics[width=0.21\linewidth]{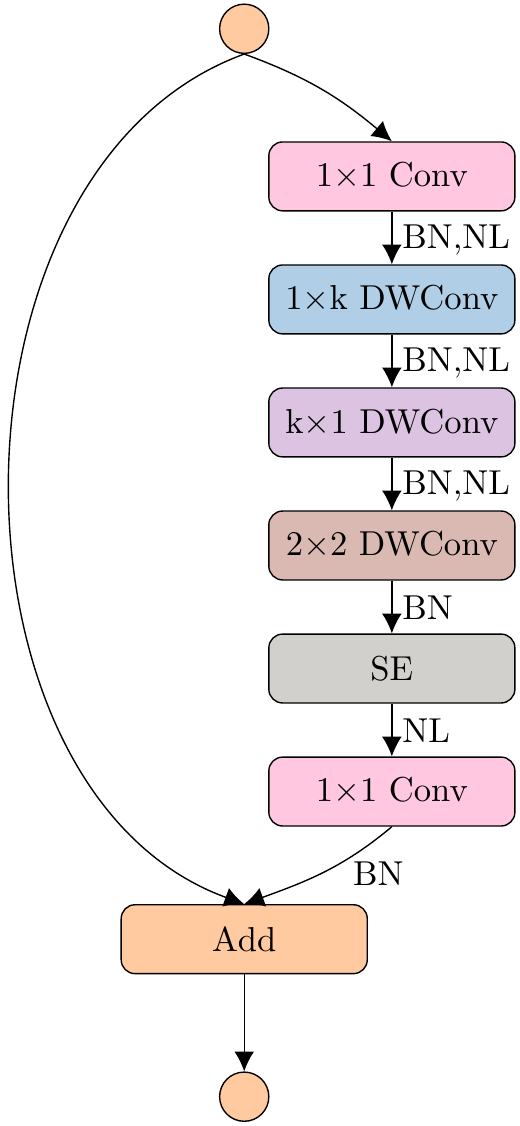}}
  \hfill
  \subfigure[]{
    \label{fig:BuildingBlock:d} %% label for second subfigure
    \includegraphics[width=0.265\linewidth]{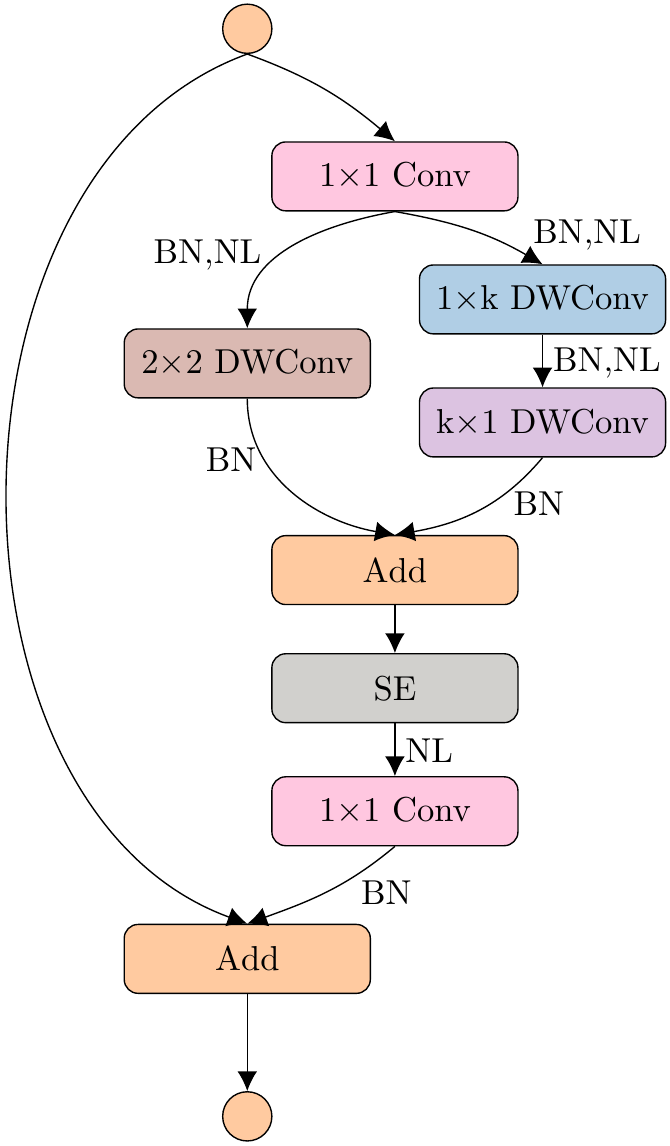}}
  \caption{Various building blocks based on the building block of MobileNetV3-Small, where the depthwise convolution is replaced with different structural designs of XSepConv: (a) standard XSepConv; (b) without $2\times2$ depthwise convolution; (c) $2\times2$ depthwise convolution placed behind spatially separated depthwise convolutions; (d) $2\times2$ depthwise convolution in parallel to spatially separated depthwise convolutions. \textbf{BN}: batch normalization \cite{ioffe2015batch}. \textbf{SE}: Squeeze-and-Excitation module \cite{hu2018squeeze}}
  \label{fig:BuildingBlock} %% label for entire figure
\end{figure}

\begin{figure}[t]
    \centering
    \includegraphics[width=0.618\linewidth]{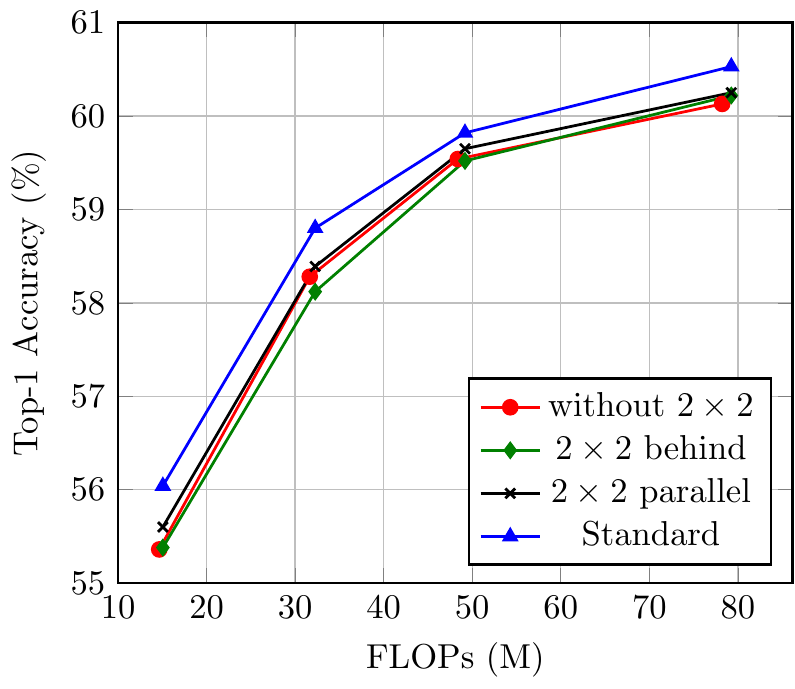}
    \caption{Trade-off curves of different structural designs in Fig.~\ref{fig:BuildingBlock} on Tiny-ImageNet}
    \label{fig:ablation}
\end{figure}

\subsubsection{Downsampling:} To evaluate the effectiveness of downsampling XSepConv as shown in Fig.~\ref{fig:XSepConv:b}, we enlarge the kernel size of the first $5\times5$ depthwise convolution of stride 2 to $7\times7$ and then replace it with downsampling XSepConv of $k=7$. %Table~\ref{table:ds7} exhibits the result, downsampling XSepConv still shows better accuracy and efficiency than downsampling depthwise convolution with large kernel size ($\ge7$). 
The trade-off curves on Tiny-ImageNet are displayed in Fig.~\ref{fig:ds7}. With fewer FLOPs, downsampling XSepConv obtains higher accuracy, proving that downsampling XSepConv indeed strikes a better trade-off between accuracy and efficiency than vanilla downsampling depthwise convolution.

\subsubsection{Larger Kernels:}Furthermore, to confirm that XSepConv can still outperform depthwise convolution with even larger kernels, we increase the kernel size of $5\times5$ depthwise convolution in the middle stage to $7\times7$ and then replace it with XSepConv of $k=7$. As shown in Fig.~\ref{fig:k7}, we observe that XSepConv gains more reductions in FLOPs as the kernel size increases from 5 to 7 compared to Fig.~\ref{fig:tradeoff:d}, but still achieves better accuracy than vanilla depthwise convolution, suggesting that XSepConv is a more efficient alternative to depthwise convolution. Interestingly, combining Fig.~\ref{fig:tradeoff:d} and Fig.~\ref{fig:k7}, we observe that both XSepConv and depthwise convolution with larger kernels exhibit higher accuracy, indicating the potentiality of larger kernels to improve model accuracy.

\subsection{Ablation Studies}\label{ablation}

In this section, we conduct a series of ablation experiments to shed light on the impact of different structural designs as shown in Fig.~\ref{fig:BuildingBlock}. We also perform ablation experiments to investigate the impact of different symmetric padding strategies.

\subsubsection{Importance of $2\times2$ Depthwise Convolution:} To assess whether $2\times2$ depthwise convolution plays an important part in boosting the performance, we remove the $2\times2$ depthwise convolution in XSepConv as shown in Fig.~\ref{fig:BuildingBlock:b}. Table~\ref{table:no2x2} illustrates the consistent decrease in accuracy on all 4 datasets although fewer parameters and less computational cost are required due to the removal of $2\times2$ depthwise convolution. For instance, removing $2\times2$ depthwise convolution results in an accuracy drop of about 0.25\% with quite little decrease in FLOPs on CIFAR-10. In order to provide an intuitive perception of the impact of the removal of $2\times2$ depthwise convolution, the trade-off curves are displayed in Fig.~\ref{fig:ablation}, clearly revealing the worse trade-off without $2\times2$ depthwise convolution. Despite reducing the computational complexity, removing $2\times2$ depthwise convolution leads to a more significant reduction in accuracy and eventually makes the trade-off worse. We also observe that the accuracy gap becomes more pronounced with fewer FLOPs. This can be attributed to the inherent structural defects of spatially separated depthwise convolutions, which will miss information that may be more important as the model becomes smaller and less redundant. Consequently, the $2\times2$ depthwise convolution in XSepConv plays an indispensable role in capturing information that may be missed by spatially separated depthwise convolutions but is important to the performance of the model.

\setlength{\tabcolsep}{4pt}
\begin{table}[t]
% \small
\begin{center}
\caption{Performance study for $2\times2$ depthwise convolution on 4 datasets}
\label{table:no2x2}
\begin{tabular}{ccccc}
\toprule\noalign{\smallskip}
XSepConv & Dataset & FLOPs & Params & Top-1 accuracy (\%)\\
\noalign{\smallskip}
\midrule
\noalign{\smallskip}
w/ $2\times2$  & \multirow{2}{*}{CIFAR-10} & 16.71M & 1.50M & \textbf{93.24}\\
w/o $2\times2$ &  & \textbf{16.45M} & \textbf{1.49M} & 92.99\\
\midrule
\noalign{\smallskip}
w/ $2\times2$  & \multirow{2}{*}{CIFAR-100} & 16.80M & 1.59M & \textbf{74.02}\\
w/o $2\times2$ &  & \textbf{16.54M} & \textbf{1.58M} & 73.89\\
\midrule
\noalign{\smallskip}
w/ $2\times2$  & \multirow{2}{*}{SVHN} & 16.71M & 1.50M & \textbf{97.97}\\
w/o $2\times2$ &  & \textbf{16.45M} & \textbf{1.49M} & 97.92\\
\midrule
\noalign{\smallskip}
w/ $2\times2$  & \multirow{2}{*}{Tiny-ImageNet} & 49.18M & 1.70M & \textbf{59.82}\\
w/o $2\times2$ &  & \textbf{48.37M} & \textbf{1.68M} & 59.62\\
\bottomrule
\end{tabular}
\end{center}
\end{table}
\setlength{\tabcolsep}{1.4pt}

\setlength{\tabcolsep}{4pt}
\begin{table}[!t]
% \small
\begin{center}
\caption{Performance comparison of different structural designs and padding strategy. FLOPs and parameters are the same and therefore not reported. ``Original-Padding'' refers to the original symmetric padding strategy proposed in \cite{wu2019convolution}}
\label{table:ablation}
\begin{tabular}{ccccc}
\toprule\noalign{\smallskip}
\multirow{2}{*}{Design} & \multicolumn{4}{c}{Top-1 Accuracy (\%) on 4 datasets}\\
\cline{2-5}
 & CIFAR-10 & CIFAR-100 & SVHN & Tiny-ImageNet\\
\noalign{\smallskip}
\midrule
\noalign{\smallskip}
XSepConv & \textbf{93.24} & \textbf{74.02} & \textbf{97.97} & \textbf{59.82} \\
XSepConv-B & 93.03 & 73.53 & 97.94 & 59.62\\
XSepConv-P & 93.06 & 73.14 & 97.91 & 59.65\\
Original-Padding & 93.05 & 73.81 & 97.94 & 58.58\\
\bottomrule
\end{tabular}
\end{center}
\end{table}
\setlength{\tabcolsep}{1.4pt}

\subsubsection{Location of $2\times2$ Depthwise Convolution:} In order to evaluate the impact of the location of $2\times2$ depthwise convolution, we consider two other variants of XSepConv: (1) XSepConv-B, where the $2\times2$ depthwise convolution is located behind spatially separated depthwise convolutions as shown in Fig.~\ref{fig:BuildingBlock:c}; (2)XSepConv-P, in which the $2\times2$ depthwise convolution is placed in parallel to spatially separated depthwise convolutions as described in Fig.~\ref{fig:BuildingBlock:d}. The performance of each variant on 4 datasets is reported in Table~\ref{table:ablation} and the trade-off curves are displayed in Fig.~\ref{fig:ablation}. We observe that XSepConv-B and XSepConv-P both result in a drop in accuracy on all 4 datasets, with the same parameters and computational cost as standard XSepConv.

Fig.~\ref{fig:ablation} shows a similar trade-off between XSepConv-B and the XSepConv without $2\times2$ depthwise convolution. The reason for the performance degradation of XSepConv-B is that the $2\times2$ depthwise convolution loses its role in capturing missing information since the information has been irreversibly partially lost after flowing through spatially separated depthwise convolutions. As a result, XSepConv-B performs similarly to the XSepConv without $2\times2$ depthwise convolution.                                                          

As for XSepConv-P, it also encounters performance degradation on all 4 datasets as illustrated in Table~\ref{table:ablation}, especially on CIFAR-100, where an accuracy decrease of 0.88\% is observed.
%the performance degradation may be caused by the mismatch of receptive fields. This problem occurs when adding the feature maps output by $2\times2$ depthwise convolution and spatially separated depthwise convolutions, because of the different receptive fields of $2\times2$ and $k\times k$. Replacing the $2\times2$ depthwise convolution with $k\times k$ depthwise convolution may help solve the problem, but it will introduce more computational cost and this is beyond the scope of this work. Besides, XSepConv-P 
The performance degradation can be attributed to the loss of the advantages of wider receptive field and greater depth. These advantages are replaced by a greater width, which in this experiment is inferior to the combination of a wider receptive field and a greater depth.

\subsubsection{Symmetric Padding Strategies:} As mentioned in Section~\ref{padding}, the original symmetric padding strategy will still encounter asymmetry in the final output of the network and we propose an improved symmetric padding strategy to overcome this shortcoming. The comparison of the two padding strategies is reported in Table~\ref{table:ablation}. Compared with original symmetric padding strategy, our proposed improved symmetric padding strategy takes the final output of the network as the major concern, thereby achieving consistent performance improvement of image classification tasks on various datasets. For example, an accuracy improvement of 0.21\% and 0.24\% are reported on CIFAR-100 and Tiny-ImageNet, respectively. In addition, it should be noticed that in this experiments, $N\%4=2$, where $N$ is the number of $2\times2$ depthwise convolution layers. This is not an ideal situation, but we can still attain stable performance improvement. Therefore, we have reason to believe that the performance improvement would be even greater in the case of $N\%4=0$.

\section{Conclusions}

In this paper, we aim to achieve a better trade-off between accuracy and efficiency for large depthwise convolutional kernels, which have shown their tendency to be employed in an increasing number of models, especially those found by NAS. To this end, we introduce XSepConv, which fuses spatially separable convolutions into depthwise convolution and adopt an extra $2\times2$ depthwise convolution coupled with improved symmetric padding strategy. 

A wide range of experiments on multiple datasets show that compared to vanilla depthwise convolution with large kernels, XSepConv attains a further decrease in both computational budget and parameter size while achieves a solid performance improvement for image classification on various datasets. Moreover, we carry out a series of ablation experiments to further verify the effectiveness of the proposed XSepConv and the improved symmetric padding strategy.

Our proposed XSepConv is a more efficient alternative to depthwise convolution and can reliably boost the performance by simply replacing depthwise convolution in models such as MobileNetV3-Small with XSepConv. Furthermore, XSepConv can be adopted in NAS to enrich the search space and develop more efficient models.

\section*{Acknowledgements}

This work is supported by the Special Foundation for the Development of Strategic Emerging Industries of Shenzhen under Grant JCYJ20170817161056260.

\bibliographystyle{splncs04}
\bibliography{egbib}
\end{document}